\documentclass[11pt]{article}

\usepackage[final]{acl}

\usepackage{times}
\usepackage{latexsym}

\usepackage[T1]{fontenc}

\usepackage[utf8]{inputenc}

\usepackage{microtype}

\usepackage{booktabs}
\usepackage{multirow} 
\usepackage{amsmath}
\usepackage{array} 
\usepackage{tcolorbox}
\usepackage[dvipsnames]{xcolor}
\usepackage{float}
\usepackage{stfloats}
\usepackage[section]{placeins}

\usepackage{soul}

\usepackage{inconsolata}

\usepackage{graphicx}
\usepackage{amssymb}

\title{SYNAPSE: Neuro-Symbolic Visual Thought-to-Text Decoding via Topological Semantic Denoising}

\author{
Akshaj Murhekar\thanks{Correspondence: \texttt{akshaj.murhekar@utexas.edu}; Abhijit Mishra: \texttt{abhijitmishra@utexas.edu}.}
\and Abhijit Mishra \\
School of Computing \\
University of Texas at Austin
}

\begin{document}
\maketitle
\begin{abstract}
Recent advances in large language models have accelerated open-vocabulary visual EEG-to-text decoding, where non-invasive neural activity recorded during visual perception is translated into natural language descriptions of viewed stimuli. However, physiological noise can introduce unrelated candidate terms, leading frozen language models to produce unstable or hallucinated descriptions. We introduce \textsc{SYNAPSE} (\textbf{Sy}mbolic \textbf{N}eural \textbf{A}lignment for \textbf{P}recise \textbf{S}emantic \textbf{E}xtraction), a lightweight neuro-symbolic framework that aims to reduce noise propagation during neural text generation through inference-time symbolic regularization. By filtering EEG-derived semantic candidates using commonsense graph structure and retrieved training exemplars, \textsc{SYNAPSE} improves semantic stability in this post-retrieval setting without end-to-end LLM fine-tuning. Experiments across two public EEG decoding benchmarks and multiple frozen LLM backends demonstrate improvements over unconstrained prompting baselines in several settings, limited degradation under object-label ablation, and competitive performance relative to selected resource-intensive fine-tuned systems, while reducing raw biometric exposure by localizing EEG processing within the encoder stack.
\end{abstract}

\section{Introduction}
\label{sec:intro}
Translating continuous neural recordings into natural language text has long remained a central objective at the intersection of computational neurotechnology and artificial intelligence. Recent advances in multi-modal generative deep learning have accelerated non-invasive brain-to-text decoding frameworks, particularly those leveraging electroencephalography (EEG) due to its temporal resolution, portability, and clinical feasibility~\cite{he2015noninvasive, benchetrit2023brain, defossez2023decoding}. In parallel, the rapid evolution of large language models (LLMs) has introduced systems capable of contextual alignment across text, vision, and speech modalities~\cite{touvron2023llama, achiam2023gpt}, driving growing interest in continuous latent translation pipelines that bridge cortical activity and natural language generation for applications in assistive communication, immersive interfaces, and neurological healthcare. In particular, recent work has focused on \textbf{visual thought-to-text translation}, where EEG activity elicited during visual perception is decoded into natural language descriptions of viewed stimuli~\cite{mishra-etal-2025-thought2text, murhekar2026sense}. Early open-vocabulary approaches relied on \textbf{end-to-end fine-tuning} of autoregressive or encoder-decoder architectures directly on multi-channel neural recordings~\cite{wang2024enhancing, lamprou2025creating, mishra-etal-2025-thought2text}; however, such tightly coupled training paradigms require large computational resources, introduce \textbf{privacy risks} through exposure of sensitive biometric signals, and often suffer from limited portability as downstream language models evolve. To address these limitations, recent decoupled frameworks such as \textsc{SENSE} replace continuous parameter optimization with \textbf{discrete latent alignment strategies} that project neural signals into a fixed vocabulary space, enabling frozen off-the-shelf LLMs to synthesize fluent text from candidate keywords in a zero-shot setting~\cite{murhekar2026sense}.

However, recent EEG-to-text systems rely primarily on learned neural representations. Because scalp EEG recordings are highly susceptible to physiological artifacts, noisy latent projections often activate semantically unrelated candidate tokens alongside target stimulus semantics, a phenomenon we term topical incongruence. When injected directly into frozen LLM prompts, these corrupted semantic cues can lead to unstable or hallucinated generation. \textbf{This motivates the question: can symbolic structure stabilize noisy neural semantic representations without retraining large language models?} To address this, we introduce \textsc{SYNAPSE}, a lightweight neuro-symbolic regularization framework that routes EEG-derived semantic candidates through a commonsense multigraph prior to decoding. A deterministic graph-filtering stage removes candidate terms that are disconnected in the induced graph, while retaining high-ranked neural candidates through a conditional union constraint. The framework further enriches decoding with grounded relational facts ($\mathbf{\mathcal{F}}$) and latent syntactic exemplars ($\mathbf{\mathcal{E}}_{\text{exemplars}}$) retrieved directly from normalized neural embeddings ($\mathbf{S} = \bar{\mathbf{x}} \cdot \bar{\mathbf{E}}^\top$), producing a filtered semantic context to reduce sensitivity to noisy candidates during generation.

We evaluate \textsc{SYNAPSE} on two public EEG-to-language benchmarks: \textsc{CVPR2017}~\cite{spampinato2017deep} and \textsc{THINGS EEG2}~\cite{gifford2022large,Wu2025UBP}, augmenting both datasets with image-grounded captions to support open-vocabulary neural text generation. Across a heterogeneous decoder suite spanning \texttt{Meta-Llama-3-8B-Instruct}, \texttt{Qwen2.5-7B-Instruct}, \texttt{GPT-4o-mini}, and \texttt{Gemini-2.5-flash-lite}, we conduct evaluation using ROUGE, BLEU, METEOR, BERTScore, an auxiliary GPT-5 LLM-as-a-judge assessment, and systematic ablation analysis. Results on \textsc{CVPR2017} show gains over unconstrained prompting and prior decoupled decoding baselines across several metrics and decoder settings, particularly under object-label ablation where symbolic regularization improves several lexical metrics. On the more challenging \textsc{THINGS EEG2} benchmark, \textsc{SYNAPSE} achieves competitive performance on several metrics under a cross-exemplar evaluation setting, while showing more consistent scores across decoder backends. Unlike compute-intensive continuous training pipelines such as \textsc{Thought2Text}~\cite{mishra-etal-2025-thought2text}, our framework operates as a lightweight inference-time layer aligned with retrieval-augmented generation and supports frozen LLM deployment. This design reduces raw biometric exposure because raw EEG signals remain localized to the encoder stack and are not used for external model adaptation. However, prompts sent to external APIs may still contain EEG-derived semantic information. To summarize, our key contributions are as follows:
\begin{itemize}
    \item We introduce \textsc{SYNAPSE}, a lightweight neuro-symbolic framework for inference-time EEG semantic regularization without end-to-end LLM fine-tuning.
    
    \item We propose a graph-filtering mechanism that removes graph-disconnected candidate terms while retaining high-ranked neural candidates.
    
    \item We develop a latent exemplar retrieval strategy that supplies syntactic templates during generation from noisy EEG-derived candidates.
    
    \item We demonstrate competitive decoding performance across multiple EEG corpora, frozen LLMs, and ablation settings.
\end{itemize}

The complete reproducibility package is available at \texttt{\url{https://github.com/akshajmurhekar/SYNAPSE-EEG}}.
\section{Related Work}
Brain-computer interfaces (BCIs) for natural language decoding have evolved from discrete command classification to continuous brain-to-text generation. Early systems primarily relied on invasive neuroprostheses to decode handwriting and speech articulation from cortical recordings~\cite{willett2023high, metzger2022generalizable}, while subsequent work extended these paradigms to non-invasive settings, demonstrating reconstruction of perceived speech~\cite{defossez2023decoding}, visual stimuli~\cite{benchetrit2023brain}, semantic narratives~\cite{tang2023semantic}, and typing-based communication~\cite{levy2025brain}. However, decoding natural language directly from EEG remains challenging due to the low spatial resolution and sensitivity to noise of scalp recordings.

Recent EEG-to-text architectures address these limitations through cross-modal alignment and self-supervised representation learning. Prior work has explored contrastive masked autoencoders~\cite{wang2024enhancing}, large-scale neural pre-training~\cite{lamprou2025creating}, interpretable semantic alignment~\cite{liu2025learning}, and semantic matching frameworks for open-vocabulary generation~\cite{tao2025see, masry2025ets}. Systems such as \textsc{Thought2Text}~\cite{mishra-etal-2025-thought2text} rely on end-to-end fine-tuning of autoregressive language models, while decoupled retrieval-based approaches such as \textsc{SENSE}~\cite{murhekar2026sense} replace model adaptation with discrete vocabulary retrieval over frozen LLMs. However, both paradigms rely primarily on connectionist representations and can remain vulnerable to physiological noise, latent representation drift, and hallucinated downstream generation. In contrast, \textsc{SYNAPSE} introduces an inference-time neuro-symbolic regularization framework that performs topological graph filtering and relational grounding over EEG-derived semantic projections, with the goal of reducing noise propagation without updating model parameters.
\section{Methodology}
\label{sec:method}

\begin{figure*}[t]
    \centering
    \includegraphics[width=\textwidth]{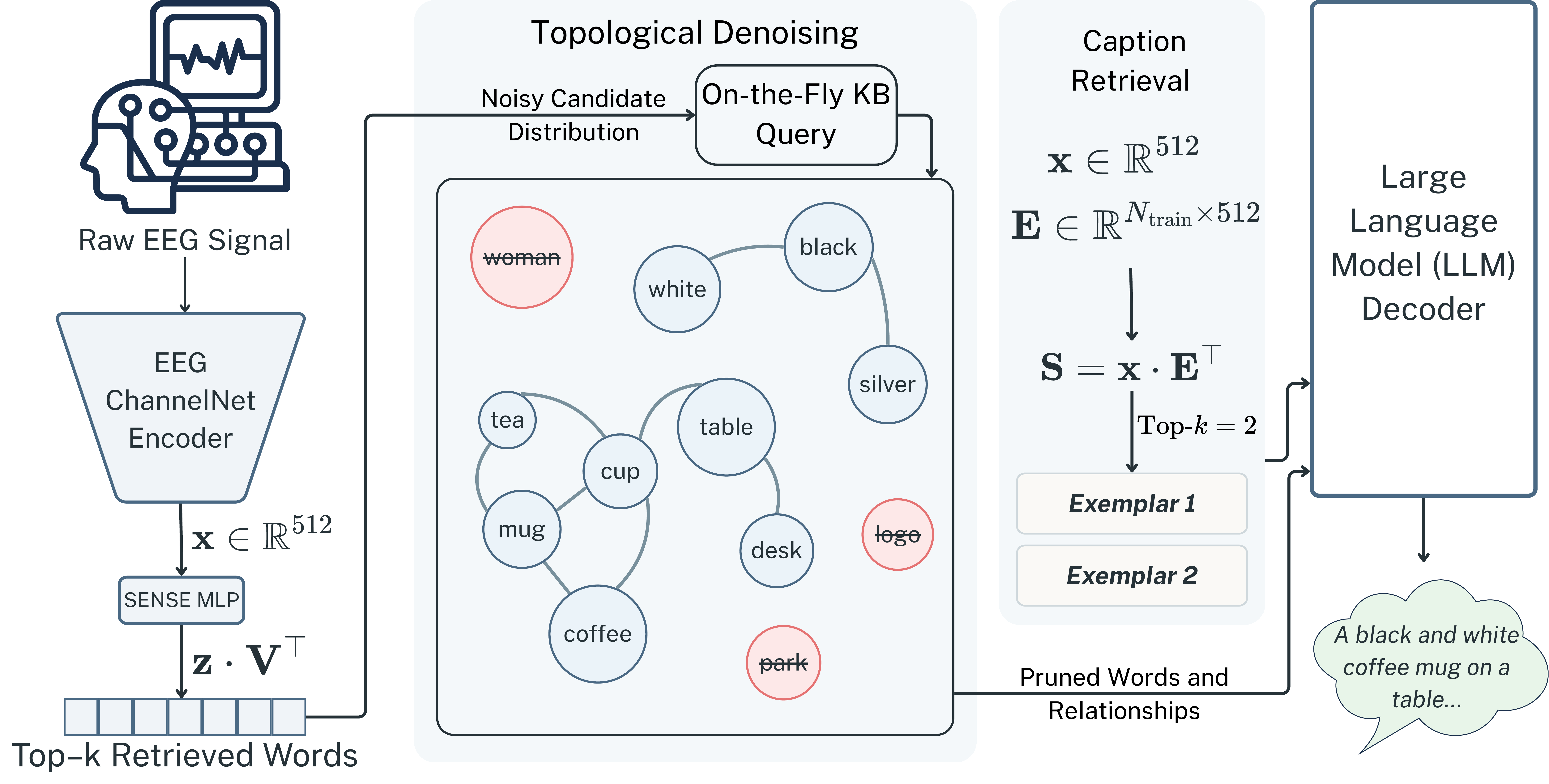}
    \caption{Overview of the \textsc{SYNAPSE} neuro-symbolic framework. Raw candidate keywords extracted from the baseline frontend via $\mathbf{z} \cdot \mathbf{V}^\top$ undergo topological graph filtering to remove graph-disconnected candidates and extract grounded relational facts. Concurrently, the unrefined neural latent vector $\mathbf{x}$ queries a historical training matrix $\mathbf{E}$ via cosine similarity $\mathbf{S} = \bar{\mathbf{x}} \cdot \bar{\mathbf{E}}^\top$ to retrieve nearest-neighbor syntactic exemplars. The {filtered} vocabulary anchors, relational context, and structural templates are fused into a consolidated prompt to guide generation across a suite of frozen large language model (LLM) decoders.}
    \label{fig:architecture}
\end{figure*}

\textsc{SYNAPSE} decodes continuous EEG recordings into natural language without requiring autoregressive LLM fine-tuning. While existing retrieval-based paradigms~\cite{murhekar2026sense} propagate unconstrained latent noise that triggers semantic drift and hallucinations, our framework introduces a {post-retrieval} neuro-symbolic intervention layer (Figure~\ref{fig:architecture}) that {filters EEG-derived semantic candidates before decoding.} This is achieved via three core operations: graph-based knowledge grounding, relational fact assertion, and non-parametric exemplar retrieval, which together provide filtered candidate terms and contextual cues to the LLM before generation.

\subsection{Problem Formulation and Neural Frontend Retrieval}

Given paired trials $\mathcal{D} = \{(X_i, s_i)\}_{i=1}^{N}$, where $X_i \in \mathbb{R}^{C \times T}$ and $s_i$ is the target caption, we adopt the frozen \textsc{SENSE} frontend~\cite{murhekar2026sense}. Raw signals $X_i$ map to language-aligned latents $\mathbf{z}_i \in \mathbb{R}^{512}$ via a ChannelNet encoder and MLP refiner ($\sim 6\text{M}$ parameters). To retrieve candidate tokens, we compute scale-normalized cosine similarity against a frozen vocabulary matrix $\mathbf{V}$. Let $\bar{\mathbf{z}}_i = \mathbf{z}_i / \|\mathbf{z}_i\|_2$ and $\bar{\mathbf{V}} = \mathbf{V}/\|\mathbf{V}\|_2$. We extract:
\begin{equation}
\mathbf{W}_{\text{raw}}
= \operatorname{TopK}\left(
\sigma(\bar{\mathbf{z}}_i \cdot \bar{\mathbf{V}}^\top),
k=15
\right).
\end{equation}

This {unfiltered} array $\mathbf{W}_{\text{raw}}$ serves as the direct operational input for our subsequent neuro-symbolic filtering stages. {We keep $k=15$ fixed across experiments to preserve reproducibility and isolate the effect of the post-retrieval symbolic layer under a common candidate budget.}

\subsection{Knowledge-Grounded Subgraph Induction and Topological Pruning}
To regularize the prediction stream via external world knowledge, \textsc{SYNAPSE} projects candidate tokens into ConceptNet~\cite{speer2017conceptnet}, {the common-sense graph evaluated in this work. The framework is decoupled from this specific graph choice and can in principle use domain-specific or personalized semantic graphs when available. We formalize the selected repository} as a weighted, directed multigraph $\mathcal{K} = (\mathcal{V}_{\mathcal{K}}, \mathcal{E}_{\mathcal{K}}, \mathcal{W}_{\mathcal{K}}, \mathcal{R}_{\mathcal{K}})$ containing vertices $\mathcal{V}_{\mathcal{K}}$, edges $\mathcal{E}_{\mathcal{K}}$, confidence weights $\mathcal{W}_{\mathcal{K}}$, and relation types $\mathcal{R}_{\mathcal{K}}$. At inference time, the $k=15$ candidate elements from $\mathbf{W}_{\text{raw}}$ are projected into the graph to construct a candidate concept vertex set $V_G \subseteq \mathcal{V}_{\mathcal{K}}$. We extract a knowledge-grounded induced subgraph $G = (V_G, E_G)$ by enforcing a minimum edge-weight threshold $w_{\text{min}} = 1.0$:
\begin{equation}
\begin{aligned}
E_G = \{\, (u,v) \in \mathcal{E}_{\mathcal{K}}
\mid{} & u,v \in V_G \\
&{}\land\; w(u,v) \geq w_{\text{min}} \,\}.
\end{aligned}
\end{equation}

To filter graph-disconnected candidates, we evaluate node connectivity within $G$. For each vertex $v \in V_G$, we compute its normalized degree centrality $C_D(v) = \deg(v)/(|V_G| - 1)$, where $\deg(v)$ is the node degree within $G$. Rather than automatically dropping all isolated nodes where $C_D(v) = 0$, \textsc{SYNAPSE} implements a hybrid selection rule utilizing a neural priority safeguard set $P = \{ w_i \in \mathbf{W}_{\text{raw}} \mid i \leq m \}$ {with $m=5$ fixed across experiments.} The final pruned vocabulary $\mathbf{W}_{\text{pruned}}$ supplied to the language decoder is defined by the conditional union:
\begin{equation}
    \mathbf{W}_{\text{pruned}} = \left\{ v \in V_G \;\middle|\; C_D(v) > 0 \;\lor\; v \in P \right\}
\end{equation}

This topological restriction applies a deterministic filtering rule while retaining high-ranked fallback candidates, {though its subtractive design remains bounded by the quality of the initial retrieved candidate set.}

\subsection{Relational Fact Extraction for Context Grounding}
To provide context for downstream generation, \textsc{SYNAPSE} traverses the edge boundaries $E_G$ of the extracted subgraph to extract available semantic assertions. This traversal is restricted to a subset of visually relevant relationship types $\mathcal{R}_{\text{meaningful}} \subset \mathcal{R}_{\mathcal{K}}$ (e.g., \textit{AtLocation, UsedFor, HasProperty, CapableOf, PartOf}). For each discovered directed edge possessing a label $r(u,v) \in \mathcal{R}_{\text{meaningful}}$, a rule-based mapping function $\tau(u, r, v)$ translates the abstract graph triple into a natural language string. The total discovered assertions are accumulated into an ordered candidate set $\mathcal{F}_{\text{all}} = \{\tau(u, r(u,v), v) \mid (u,v) \in E_G \land r(u,v) \in \mathcal{R}_{\text{meaningful}}\}$. To handle variable relational density and maintain prompt efficiency, we slice this sequence to a {fixed} capacity threshold $N_{\text{facts}} = 5$, {yielding} the context set $\mathcal{F}$:
\begin{equation}
    \mathcal{F} = \mathcal{F}_{\text{all}}[1 : N_{\text{facts}}]
\end{equation}
Including $\mathcal{F}$ in the final generation prompt provides the downstream decoder with explicit commonsense context from the induced graph.

\subsection{Cross-Modal Exemplar Retrieval for Syntactic Context}

To supply the decoder with syntactic context without autoregressive fine-tuning, \textsc{SYNAPSE} implements a non-parametric, cross-modal exemplar retrieval strategy executed directly on the unrefined neural latent space. Let the historical training split index be represented by the frozen matrix $\mathbf{E} \in \mathbb{R}^{N_{\text{train}} \times 512}$ containing the stacked ChannelNet embeddings of historical trials. During inference, the active unrefined feature vector $\mathbf{x} \in \mathbb{R}^{512}$ serves as the continuous query. Let $\bar{\mathbf{x}} = \mathbf{x}/\|\mathbf{x}\|_2$ and let $\bar{\mathbf{E}}$ denote the row-wise $L_2$-normalized training matrix. We compute the cosine similarity score vector $\mathbf{S} \in \mathbb{R}^{N_{\text{train}}}$ as:
\begin{equation}
    \mathbf{S} = \bar{\mathbf{x}} \cdot \bar{\mathbf{E}}^\top
\end{equation}
We isolate the indices of the nearest-neighbors using $\text{Top-k} = 2$. Their corresponding training captions are retrieved {strictly from the historical training partition to construct the syntactic exemplar set $\mathcal{E}_{\text{exemplars}}$. These exemplars are intended to provide syntactic templates rather than new supervision at inference time; however, when train and test splits share concepts, retrieval can still carry concept-level information, which we discuss as a limitation.}

\subsection{Linguistic Decoding Loop}

The final phase consolidates four data streams into a structured prompt: (i) the graph-filtered token sequence $W_{\text{pruned}}$, (ii) grounded semantic facts $\mathcal{F}$, (iii) cross-modal syntactic exemplars $\mathcal{E}_{\text{exemplars}}$, and (iv) a target visual object class anchor with fixed generation constraints. The prompt is routed through a frozen, off-the-shelf LLM decoder; {in our evaluated setup, raw EEG processing remains local, while decoder calls may use external API backends depending on the selected model.}
\section{Experiments}
We evaluate \textsc{SYNAPSE} on two EEG-to-text benchmarks using multiple frozen language decoders and complementary translation metrics.

\subsection{Experimental Setup and Datasets}
\paragraph{\textsc{Thought2Text} / \textsc{SENSE} Benchmark:} We evaluate \textsc{SYNAPSE} on the multi-modal EEG-to-text benchmark corpus from~\cite{mishra-etal-2025-thought2text,murhekar2026sense}. This dataset contains non-invasive EEG recordings from six healthy subjects viewing $2,000$ ImageNet visual stimuli, paired with natural language captions. Accounting for multiple subject trials, the corpus contains $9,940$ samples, partitioned into $7,959$ training samples and $1,987$ test samples. Partitioning is performed at the trial level, so subjects are shared across train and test splits; the validation split is not used in this work. Following \textsc{SENSE} setup, raw neural features are encoded into $512$-dimensional vector spaces via a frozen ChannelNet architecture using the default pipeline hyperparameters established in Section \ref{sec:method}.

\paragraph{THINGS EEG2 Cross-Exemplar Suite:} For larger-scale evaluation, we evaluate \textsc{SYNAPSE} on the public THINGS EEG2 repository~\cite{gifford2022large}. The dataset covers $1,654$ distinct visual concepts. Because the original corpus lacks target textual descriptors, we adopt the data-bootstrapping paradigm introduced in \textsc{Thought2Text} ~\cite{mishra-etal-2025-thought2text} and use a large language model, GPT-4-Omni ~\cite{achiam2023gpt} to generate reference text captions for each visual stimulus across all $1,654$ concepts. Our experiments use the first five healthy subjects. The training split contains $8,270$ samples, corresponding to 5 image exemplars per concept, and the test split contains $1,654$ samples, corresponding to the 6th unseen visual exemplar for each concept. For both splits, deterministic cross-trial averaging is executed over the 4 repetitive presentation epochs to improve the signal-to-noise ratio (SNR). This protocol evaluates cross-exemplar generalization to unseen images of seen concepts, rather than concept-disjoint generalization to entirely unseen concepts.

The following prompt template was used to construct reference captions for image stimuli lacking textual annotations:
\begin{figure}[H]
\centering
\begin{tcolorbox}[colback=gray!10, colframe=gray!30, coltitle=black,
    fonttitle=\bfseries, left=0.1cm, right=0.1cm, bottom=-1mm-0.5mm]
\{"role": "system", "content": "You are a helpful assistant."\},\\
\{"role": "user", "content": "<image> <object\_string> Describe this in one sentence:"\}
\tcbsubtitle[halign=center]{Input prompt}
\end{tcolorbox}
\end{figure}

Because the reference captions are generated through a language-model-assisted protocol, our results are best interpreted as controlled comparisons within the established \textsc{Thought2Text}/\textsc{SENSE} setup rather than direct measurements of human-written descriptions. We reuse the released preprocessing, channel handling, and feature extraction pipeline from these prior works; our contribution begins from the frozen neural feature coordinates and candidate-token retrieval stage.

\subsection{Evaluated Decoders and Metrics}
We route consolidated prompts through open-weights decoders---\texttt{Qwen2.5-7B-Instruct}, \texttt{Meta-LLaMA-3-8B-Instruct}---and commercial endpoints---\texttt{GPT-4o-mini} and \texttt{Gemini-2.5-flash-lite}. Following \textsc{Thought2Text}~\cite{mishra-etal-2025-thought2text} and prior baselines~\cite{murhekar2026sense}, we evaluate translation fidelity with BLEU-1/4~\cite{papineni2002bleu}, ROUGE-1/2/L~\cite{lin2004rouge}, METEOR~\cite{banerjee2005meteor}, and BERTScore~\cite{zhang2019bertscore}. We additionally report GPT-5-mini~\cite{singh2025openaigpt5card} fluency and adequacy judgments on a standardized 1--5 scale as auxiliary diagnostics rather than primary comparison evidence.

\section{Results and Discussion}

We evaluate \textsc{SYNAPSE} across multiple decoder backends, contrasting open-weights architectures with commercial models accessed through API endpoints. Translation quality is analyzed using lexical overlap and embedding-similarity metrics, together with ablations designed to isolate the contribution of each neuro-symbolic component. The results show that graph filtering, relational grounding, and exemplar retrieval improve several decoding settings, while also revealing decoder-specific trade-offs and limitations in adequacy.

\begin{table*}[!t]
    \centering
    \small
    \setlength{\tabcolsep}{3.8pt}
    \begin{tabular}{ll ccc ccc cc | cc} 
         \toprule
         \multicolumn{2}{l}{\textbf{Ablation Configuration}} & \multicolumn{2}{c}{\textbf{ROUGE}} & \textbf{R-L} & \multicolumn{2}{c}{\textbf{BLEU}} & \textbf{MET.} & \textbf{BERT} & \multicolumn{2}{c}{\textbf{GPT-5 Evaluation}}\\ 
         \textbf{ID} & \textbf{Decoder Model} & \textbf{R-1} & \textbf{R-2} & & \textbf{B-1} & \textbf{B-4} & & \textbf{Score} & \textbf{Fluency} & \textbf{Adequacy} \\ 
         \midrule
         $\mathcal{A}_1$ & $\mathrm{GPT\textrm{-}4o\textrm{-}mini}$ & 31.7 & 8.8 & 28.1 & \underline{\textit{27.6}} & 5.9 & \underline{\textbf{28.2}} & 90.2 & \textit{4.79} & 1.37 \\
         \scriptsize{(Full)} & $\mathrm{Gemini~2.5~Flash~Lite}$ & 32.3 & 8.6 & 28.6 & 27.3 & 5.8 & 27.5 & 90.3 & 4.68 & 1.37 \\
          & $\mathrm{LLaMA\textrm{-}3\textrm{-}8B}$ & 30.0 & 8.2 & 27.2 & 25.1 & 5.4 & 26.8 & 90.2 & 4.71 & 1.35 \\
          & $\mathrm{Qwen2.5\textrm{-}7B}$ & 32.2 & \underline{\textbf{9.3}} & 28.7 & 27.3 & \underline{\textit{6.1}} & 28.0 & \underline{\textit{90.4}} & 4.48 & 1.35 \\
         \midrule
         $\mathcal{A}_2$ & $\mathrm{GPT\textrm{-}4o\textrm{-}mini}$ & 30.8 & 8.2 & 27.3 & 26.9 & 5.5 & 27.0 & 90.1 & 4.78 & 1.33 \\
         \scriptsize{($m = 0$)} & $\mathrm{Gemini~2.5~Flash~Lite}$ & 31.4 & 7.6 & 27.7 & 26.5 & 5.2 & 25.5 & 90.1 & 4.61 & 1.31 \\
          & $\mathrm{LLaMA\textrm{-}3\textrm{-}8B}$ & 29.5 & 7.9 & 26.8 & 24.6 & 5.2 & 26.1 & 90.1 & 4.69 & 1.33 \\
          & $\mathrm{Qwen2.5\textrm{-}7B}$ & 31.6 & 8.4 & 28.0 & 26.7 & 5.6 & 26.7 & 90.3 & 4.44 & 1.31 \\
         \midrule
         $\mathcal{A}_3$ & $\mathrm{GPT\textrm{-}4o\textrm{-}mini}$ & 31.6 & 8.7 & 27.9 & \underline{\textit{27.6}} & 5.9 & \underline{\textbf{28.2}} & 90.2 & 4.78 & 1.36 \\
         \scriptsize{($-$Obj)} & $\mathrm{Gemini~2.5~Flash~Lite}$ & 32.3 & 8.4 & 28.4 & 27.4 & 5.7 & 27.4 & 90.3 & 4.64 & 1.35 \\
          & $\mathrm{LLaMA\textrm{-}3\textrm{-}8B}$ & 30.4 & 8.6 & 27.3 & 25.8 & 5.5 & 27.4 & 90.2 & 4.71 & 1.34 \\
          & $\mathrm{Qwen2.5\textrm{-}7B}$ & 32.3 & \underline{\textit{9.2}} & 28.6 & \underline{\textit{27.6}} & \underline{\textit{6.1}} & \underline{\textbf{28.2}} & 90.3 & 4.38 & 1.34 \\
         \midrule
         $\mathcal{A}_4$ & $\mathrm{GPT\textrm{-}4o\textrm{-}mini}$ & 31.6 & 8.9 & 28.1 & \underline{\textit{27.6}} & 6.0 & \underline{\textbf{28.2}} & 90.2 & 4.77 & 1.36 \\
         \scriptsize{($-$Exemp)} & $\mathrm{Gemini~2.5~Flash~Lite}$ & \textbf{32.7} & 8.7 & \textbf{28.9} & \underline{\textbf{27.8}} & 5.9 & 27.9 & 90.3 & 4.67 & 1.36 \\
          & $\mathrm{LLaMA\textrm{-}3\textrm{-}8B}$ & 29.8 & 8.2 & 27.1 & 25.0 & 5.4 & 26.7 & 90.2 & 4.72 & 1.36 \\
          & $\mathrm{Qwen2.5\textrm{-}7B}$ & 32.2 & \underline{\textbf{9.3}} & \textit{28.8} & 27.3 & \underline{\textbf{6.2}} & 27.9 & \underline{\textit{90.4}} & 4.46 & 1.35 \\
         \midrule
         $\mathcal{A}_5$ & $\mathrm{GPT\textrm{-}4o\textrm{-}mini}$ & 30.9 & 8.2 & 27.3 & 26.9 & 5.5 & 28.0 & 90.0 & 4.78 & 1.35 \\
         \scriptsize{($-$Fac)} & $\mathrm{Gemini~2.5~Flash~Lite}$ & 32.3 & 8.4 & 28.3 & \underline{\textbf{27.8}} & 5.6 & 28.0 & 90.2 & 4.66 & 1.34 \\
          & $\mathrm{LLaMA\textrm{-}3\textrm{-}8B}$ & 29.1 & 8.1 & 26.6 & 24.0 & 5.3 & 26.2 & 90.1 & 4.72 & 1.38 \\
          & $\mathrm{Qwen2.5\textrm{-}7B}$ & 32.0 & \underline{\textit{9.2}} & 28.6 & 26.8 & \underline{\textbf{6.2}} & 26.9 & \textbf{90.5} & 4.57 & 1.36 \\
         \midrule
         $\mathcal{A}_6$ & $\mathrm{GPT\textrm{-}4o\textrm{-}mini}$ & 28.0 & 5.9 & 24.3 & 24.1 & 4.1 & 26.2 & 89.4 & 4.78 & 1.31 \\
         \scriptsize{(Base)} & $\mathrm{Gemini~2.5~Flash~Lite}$ & 31.0 & 7.5 & 26.9 & 26.6 & 5.0 & 27.5 & 89.8 & 4.58 & 1.30 \\
          & $\mathrm{LLaMA\textrm{-}3\textrm{-}8B}$ & 25.3 & 5.3 & 21.6 & 21.4 & 3.5 & 24.8 & 89.1 & 4.46 & 1.25 \\
          & $\mathrm{Qwen2.5\textrm{-}7B}$ & 29.2 & 6.6 & 24.9 & 24.9 & 4.4 & 26.0 & 89.6 & 4.24 & 1.27 \\
          \midrule
        $\mathcal{A}_7$ & $\mathrm{GPT\textrm{-}4o\textrm{-}mini}$ & 20.9 & 2.9 & 19.1 & 18.4 & 2.9 & 18.0 & 88.6 & 4.60 & 1.09 \\
        \scriptsize{(Random)} & $\mathrm{Gemini~2.5~Flash~Lite}$ & 21.3 & 3.1 & 19.9 & 16.1 & 2.7 & 16.2 & 88.7 & 4.11 & 1.06 \\
         & $\mathrm{LLaMA\textrm{-}3\textrm{-}8B}$ & 19.4 & 2.8 & 17.9 & 15.4 & 2.5 & 17.5 & 88.1 & 4.23 & 1.08 \\
         & $\mathrm{Qwen2.5\textrm{-}7B}$ & 23.1 & 3.5 & 21.0 & 19.3 & 3.0 & 18.6 & 88.7 & 4.08 & 1.07 \\
         \midrule
         \multicolumn{11}{l}{\textbf{Baselines \& Comparative Frameworks}} \\
         \midrule
         \textsc{SENSE} & $\mathrm{GPT\textrm{-}4o\textrm{-}mini}$$_{\text{{With Obj}}}$ & 30.6 & 8.2 & 28.2 & 24.9 & 5.6 & 25.3 & 89.8 & 4.75 & \underline{\textit{1.40}} \\
          & $\mathrm{Gemini~2.5~Flash~Lite}$$_{\text{{With Obj}}}$ & 31.5 & 8.5 & 28.7 & 25.2 & 5.6 & 26.1 & 89.7 & 4.77 & \underline{\textit{1.40}} \\
          & $\mathrm{LLaMA\textrm{-}3\textrm{-}8B}$$_{\text{{With Obj}}}$ & 12.6 & 2.2 & 11.6 & 6.8 & 1.5 & 12.5 & 88.1 & 4.56 & 1.33 \\
          & $\mathrm{Qwen2.5\textrm{-}7B}$$_{\text{{With Obj}}}$ & 21.8 & 5.9 & 19.9 & 13.8 & 3.2 & 15.0 & 87.6 & 3.47 & 1.33 \\
         \midrule
         \textsc{Thought2Text} & $\mathrm{LLaMA\textrm{-}3\textrm{-}8B}$$_{\text{{ALL}}}$ & 30.0 & 8.1 & 26.6 & 25.5 & 5.5 & 26.3 & 89.0 & \textbf{4.82} & \textbf{1.58} \\
          & $\mathrm{Qwen2.5\textrm{-}7B}$$_{\text{{ALL}}}$ & 26.4 & 4.6 & 22.8 & 22.7 & 3.7 & 21.1 & 88.0 & 4.75 & 1.28 \\
         
         \bottomrule
    \end{tabular}
    \caption{Ablation tracking and comparative framework performance over the ImageNet-EEG benchmark. Higher values indicate superior performance across all metrics. Top results are \textbf{bolded} (tied \underline{\textbf{underlined}}) and second-best are \textit{italicized} (tied \underline{\textit{underlined}}). Setup configurations: $\mathcal{A}_1$: Full framework; $\mathcal{A}_2$: $m=0$; $\mathcal{A}_3$: Full w/o Object Labels; $\mathcal{A}_4$: Full w/o Exemplars; $\mathcal{A}_5$: Full w/o Facts; $\mathcal{A}_6$: Minimal baseline (BoW + Obj only); $\mathcal{A}_7$: Randomized prompt-control baseline with one true word retained, 14 random vocabulary tokens, and random training captions.}
    \label{tab:quant_metrics}
\end{table*}
\FloatBarrier

\subsection{Quantitative Performance and Component Ablations}

Under full configuration ($\mathcal{A}_1$), \textsc{SYNAPSE} achieves competitive performance across the evaluated decoders by combining filtered vocabulary vectors $\mathbf{W}_{\text{pruned}}$, relational knowledge blocks $\mathbf{\mathcal{F}}$, and cross-modal syntactic templates $\mathbf{\mathcal{E}}_{\text{exemplars}}$. A key pattern emerges when evaluating the framework without the explicit object-label anchor ($\mathcal{A}_3$, \textit{w/o Perceptual Object Anchors}). In retrieval based pipelines like \textsc{SENSE} \cite{murhekar2026sense}, withholding visual class labels can reduce translation quality, suggesting reliance on the object-label signal. Conversely, under \textsc{SYNAPSE}, withholding explicit object labels yields little performance degradation; e.g., \texttt{Qwen2.5-7B} maintains a stable ROUGE-1 of $32.3$ and BLEU-4 of $6.1$. Across decoders, withholding explicit object labels causes no meaningful degradation across the reported lexical metrics and in some cases improves performance, suggesting that the symbolic context reduces reliance on the explicit object-label anchor.

Component ablations show that most structured \textsc{SYNAPSE} variants remain competitive, while more aggressive controls degrade more clearly. Removing relational facts ($\mathcal{A}_5$, \textit{w/o Relational Context Facts}) produces modest lexical drops for several decoders, but the effect is not uniform: Gemini and Qwen retain strong scores on selected metrics, indicating that some decoder backends can partially compensate using the filtered BoW and exemplar context alone. Similarly, removing cross-modal exemplars ($\mathcal{A}_4$, \textit{w/o In-Context Exemplars}) does not uniformly harm performance; it matches or slightly improves several lexical metrics for GPT-4o-mini, Gemini, and Qwen, while producing only small changes for LLaMA. These patterns suggest that the structured components are useful but partially redundant, with their marginal contribution depending on the downstream decoder. In contrast, the minimal configuration ($\mathcal{A}_6$, \textit{Clean BoW Only}) shows clearer degradation, especially for LLaMA and Qwen fluency/lexical scores, indicating that relational and exemplar context help stabilize generation beyond filtered keywords alone.

Because several ablation scores are numerically close, we further evaluate whether the top configuration within each decoder track differs significantly from its runner-up (Table~\ref{tab:significance}). We use 10,000 bootstrap resamples to estimate 95\% confidence intervals and paired Wilcoxon signed-rank tests for paired top-vs-runner-up comparisons.

\begin{table}[H]
\centering
\footnotesize
\setlength{\tabcolsep}{2.5pt}
\begin{tabular}{llcc}
\toprule
\textbf{Dec.} & \textbf{Met.} & \textbf{Best config. + 95\% CI} & \textbf{$p$} \\
\midrule
LLaMA & R-1  & 30.4 [29.8, 31.0] ($-$Obj)   & $.0182*$ \\
LLaMA & B-1  & 25.8 [25.2, 26.4] ($-$Obj)   & $.0018*$ \\
LLaMA & B-4  & 5.5 [5.2, 5.9] ($-$Obj)      & $.0156*$ \\
LLaMA & MET. & 27.4 [26.7, 28.1] ($-$Obj)   & $.0031*$ \\
Gemini & R-1 & 32.7 [32.0, 33.4] ($-$Exemp) & $.0045*$ \\
Gemini & B-4 & 5.9 [5.6, 6.2] ($-$Exemp)    & $.0064*$ \\
\bottomrule
\end{tabular}
\caption{Significant top-vs-runner-up ablation comparisons. Non-significant tracks and metrics are omitted.}
\label{tab:significance}
\end{table}

Table~\ref{tab:significance} reinforces the decoder-specific nature of the ablation results. For LLaMA-3-8B, the no-object-label configuration ($\mathcal{A}_3$) significantly improves several lexical metrics over the runner-up, supporting the interpretation that \textsc{SYNAPSE} reduces reliance on explicit visual object anchors for this decoder. For Gemini-2.5-Flash, the no-exemplar configuration ($\mathcal{A}_4$) significantly improves ROUGE-1 and BLEU-4, suggesting that exemplar retrieval can occasionally impose stylistic constraints rather than uniformly helping. For GPT-4o-mini and Qwen2.5-7B, as well as the remaining LLaMA and Gemini metrics, the best and runner-up configurations are statistical ties under this analysis.

The randomized prompt-control baseline ($\mathcal{A}_7$) further separates prompt formatting effects from informative neural retrieval. This setting preserves the same prompt template but retains only one true candidate word, replaces the remaining $14$ slots with random vocabulary tokens, and substitutes nearest-neighbor exemplars with random training captions. Performance drops sharply across all decoders; for example, GPT-4o-mini falls from $31.7$ to $20.9$ ROUGE-1 and from $5.9$ to $2.9$ BLEU-4. This indicates that the gains of the structured configurations are not explained by prompt format alone, but depend on preserving informative EEG-derived candidate words and relevant exemplar context.

\subsection{Characterizing Topological Incongruence and Pruning Dynamics}

To isolate the effect of graph filtering before language generation, we first measure candidate keyword quality before and after the topological pruning stage (Table~\ref{tab:retrieval_quality}). Because this filter is strictly subtractive, it cannot recover concepts missing from the initial EEG-to-word retrieval set; instead, it can only remove candidates judged structurally disconnected from the induced ConceptNet subgraph.

\begin{table}[H]
\centering
\small
\begin{tabular}{lccc}
\toprule
\textbf{Phase} & \textbf{Precision (\%)} & \textbf{Recall (\%)} & \textbf{Dropped} \\
\midrule
Pre-Graph & 14.58 & 20.44 & -- \\
Post-Graph & 16.92 & 17.38 & 3.87 \\
Delta & +2.34 & -3.06 & 25.8\% \\
\bottomrule
\end{tabular}
\caption{Candidate keyword quality on the ImageNet-EEG test split.}
\label{tab:retrieval_quality}
\end{table}

The retrieval analysis shows the expected precision-recall trade-off. Graph filtering raises mean candidate precision from $14.58\%$ to $16.92\%$ while reducing recall from $20.44\%$ to $17.38\%$, removing an average of $3.87$ words per sample. This supports the intended role of the graph layer as a noise-reduction mechanism rather than a recall-improving retrieval module.

Filtering singletons ($C_D(\mathbf{v}) = 0$) generated by \textit{topical incongruence} can reduce the number of disconnected candidates passed to the decoder. Parametric sensitivity sweeps over the priority safeguard hyperparameter $m \in \{0, 5\}$ across both corpora show distinct pruning trends (Table~\ref{tab:pruning_stats}). 

Our default configuration ($m=5$) provides a practical balance on ImageNet-EEG, retaining a mean pool of $11.13$ retained terms while filtering $3.87$ words per trial (a $25.8\%$ macro pruning rate). Removing the safeguard ($m=0$) increases macro dataset pruning to $33.5\%$, removing up to 13 words in some trials. While the unconstrained filter removes more candidate words, it can also remove valid visual descriptors that lack explicit common-sense paths within static databases.

Conversely, the connectivity constraint surfaces a vulnerability to \textit{spurious relational co-activation}. If the connectionist frontend retrieves multiple incongruent candidate words that share an explicit link in the knowledge base, they can form an isolated subgraph. Because their degree metrics remain non-zero ($C_D(\mathbf{v}) > 0$), these terms can bypass topological pruning and remain in the prompt context. While our non-parametric blueprints $\mathbf{\mathcal{E}}_{\text{exemplars}}$ and explicit facts set $\mathbf{\mathcal{F}}$ may reduce the effect of these remaining candidates, this artifact shows that a permissive connectivity constraint retains rare but potentially valid candidates at the cost of allowing clustered token noise to occasionally evade filtration.

Qualitative analysis (Appendix~\ref{sec:appendix_qualitative}) illustrates both successful filtering and failure cases. Table~\ref{tab:qualitative} reports the refined, pruned BoW inputs perceived by the decoder. Some examples show that the system removes discordant candidates through this dual-stage process: when the frontend emits isolated activations such as \textit{daisy} and \textit{flower} for a target \textit{mushroom}, the graph filter removes these structural outliers. By providing the LLM with the remaining filtered tokens alongside grounded relational facts ($\mathbf{\mathcal{F}}$), the decoder can produce a description closer to the target despite residual noise. Other cases show the limits of this approach: when related but incorrect candidates such as \textit{pizza}, \textit{pepperoni}, and \textit{cheese} remain after filtering for a \textit{flower} stimulus, the decoder must resolve a coherent but semantically incorrect prompt context. These cases underscore the framework's reliance on frontend representation quality.

\begin{table*}[t]
\centering
\small
\setlength{\tabcolsep}{12pt}
\renewcommand{\arraystretch}{1.15}
\begin{tabular}{l cc c}
\toprule
 & \multicolumn{2}{c}{\textbf{ImageNet-EEG}} & \textbf{Things EEG2} \\
\cmidrule(lr){2-3} \cmidrule(lr){4-4}
\textbf{Topological Filtration Metric} & \textbf{$m=5$ (Default)} & \textbf{$m=0$ (None)} & $m=5$ \\
\midrule
Average Retained Word Pool (words)        & 11.13 & 9.98  & 11.41 \\
Average Dropped per Sample (words)       & 3.87  & 5.02  & 3.59  \\
Macro Dataset Pruning Rate (\%)           & 25.8  & 33.5  & 23.9  \\
Maximum Dropped in a Single Trial (words) & 10    & 13    & 8     \\
\bottomrule
\end{tabular}
\caption{Macro-level topological pruning and error-correction sensitivity statistics compared across the ImageNet-EEG ($N=1,987$) and Things EEG2 ($N=1,654$) datasets under variable priority safeguard constraints. For all evaluations, the minimum tokens dropped in a single trial was 0.}
\label{tab:pruning_stats}
\end{table*}

\paragraph{Latency Overhead.}
We also measure the per-trial runtime overhead of the inference-time layer under restricted resources of approximately 8GB RAM. The full pipeline, including EEG embedding, SENSE token retrieval, ConceptNet filtering, and exemplar retrieval, requires 537 ms per trial. This suggests that the symbolic layer introduces a modest practical overhead relative to the frozen frontend and decoder pipeline, although deployment latency will also depend on the selected LLM backend.

\subsection{Neuro-Symbolic Decoupling vs. Fine-Tuning}
End-to-end connectionist architectures optimize language-model behavior directly from continuous neural features, which requires the decoder to learn from noisy neural representations during training. Our results suggest that this can be difficult under volatile, non-invasive recordings, where retrieval noise can propagate into the generated text.

This issue is more visible within compact open-weights models (\texttt{Meta-LLaMA-3-8B} and \texttt{Qwen2.5-7B}). These decoders show larger drops when given noisier retrieval inputs. In this setting, the decoder must generate fluent text while also resolving contradictory candidate terms, making outputs more sensitive to retrieval errors.

\textsc{SYNAPSE} addresses this issue by separating neural retrieval from symbolic filtering. By applying deterministic graph filtering before prompt construction, the framework removes graph-disconnected candidates before they are passed to the decoder. The downstream model then receives a more constrained prompt containing the filtered token set $\mathbf{W}_{\text{pruned}}$, explicit prose assertions $\mathbf{\mathcal{F}}$, and nearest-neighbor syntactic exemplars $\mathbf{\mathcal{E}}_{\text{exemplars}}$. This design enables lightweight open-weights models to match or exceed fine-tuned baselines on selected lexical metrics while still trailing them on some adequacy judgments.

\begin{table*}[t]
    \centering
    \small
    \setlength{\tabcolsep}{3.8pt}
    \begin{tabular*}{\textwidth}{l@{\extracolsep{\fill}}l ccc ccc cc | cc} 
         \toprule
         \multicolumn{2}{l}{\textbf{Ablation Configuration}} & \multicolumn{2}{c}{\textbf{ROUGE}} & \textbf{R-L} & \multicolumn{2}{c}{\textbf{BLEU}} & \textbf{MET.} & \textbf{BERT} & \multicolumn{2}{c}{\textbf{GPT-5 Evaluation}}\\ 
         \textbf{ID} & \textbf{Decoder Model} & \textbf{R-1} & \textbf{R-2} & & \textbf{B-1} & \textbf{B-4} & & \textbf{Score} & \textbf{Fluency} & \textbf{Adequacy} \\ 
         \midrule
         $\mathcal{B}_1$ & $\mathrm{GPT\textrm{-}4o\textrm{-}mini}$ & 21.4 & 2.5 & 18.3 & \textit{19.3} & 2.7 & 16.5 & 87.4 & \underline{\textbf{4.8}} & 1.0 \\
         \scriptsize{(Full)} & $\mathrm{Gemini~2.5~Flash~Lite}$ & 21.2 & 2.8 & 18.2 & 17.6 & 2.7 & 15.9 & 87.4 & 4.4 & 1.0 \\
          & $\mathrm{LLaMA\textrm{-}3\textrm{-}8B}$ & 20.3 & \textit{3.9} & 18.3 & 17.8 & \textit{3.0} & \textit{17.1} & 87.4 & 4.5 & 1.0 \\
          & $\mathrm{Qwen2.5\textrm{-}7B}$ & \textit{21.8} & 3.0 & 18.9 & \underline{\textbf{19.4}} & 2.9 & 16.7 & 87.4 & 4.2 & 1.0 \\
         \midrule
         $\mathcal{B}_2$ & $\mathrm{GPT\textrm{-}4o\textrm{-}mini}$ & 21.1 & 2.1 & 18.1 & 18.4 & 2.5 & 15.9 & 87.4 & \underline{\textit{4.7}} & 1.0 \\
         \scriptsize{($-$Exemp)} & $\mathrm{Gemini~2.5~Flash~Lite}$ & 21.4 & 2.4 & 18.4 & 18.5 & 2.7 & 16.3 & 87.2 & 4.3 & 1.0 \\
          & $\mathrm{LLaMA\textrm{-}3\textrm{-}8B}$ & 15.2 & 1.8 & 12.4 & 13.3 & 2.1 & 14.5 & 87.0 & 4.2 & 1.0 \\
          & $\mathrm{Qwen2.5\textrm{-}7B}$ & 21.5 & 2.5 & 18.1 & \underline{\textbf{19.4}} & 2.7 & 16.8 & 87.0 & 3.9 & 1.0 \\
         \midrule
         \multicolumn{11}{l}{\textbf{Baselines \& Comparative Frameworks}} \\
         \midrule
         \textsc{SENSE} & $\mathrm{GPT\textrm{-}4o\textrm{-}mini}$ & 21.2 & 2.7 & \textit{19.4} & 13.4 & 2.1 & 14.1 & \textbf{88.5} & \underline{\textbf{4.8}} & 1.0 \\
         \scriptsize{(Base)} & $\mathrm{Gemini~2.5~Flash~Lite}$ & \textbf{22.5} & \underline{\textbf{4.2}} & \textbf{21.2} & 13.2 & 2.4 & 15.1 & \textit{88.2} & \underline{\textit{4.7}} & 1.0 \\
          & $\mathrm{LLaMA\textrm{-}3\textrm{-}8B}$ & 19.9 & \underline{\textbf{4.2}} & 17.1 & 17.5 & \textbf{3.1} & \textbf{17.8} & 86.2 & 4.1 & 1.0 \\
          & $\mathrm{Qwen2.5\textrm{-}7B}$ & 9.4 & 1.2 & 8.5 & 3.6 & 0.7 & 4.3 & 84.7 & 2.0 & 1.0 \\
         \bottomrule
    \end{tabular*}
    \caption{Ablation tracking performance over the THINGS EEG2 cross-exemplar evaluation suite. Higher values indicate superior performance across all metrics. Setup configurations are explicitly mapped as follows: $\mathcal{B}_1$: Full framework (BoW + Obj + Facts + Exemp); $\mathcal{B}_2$: Full framework w/o Exemplars.}
    \label{tab:things_metrics}
\end{table*}

\subsection{Scaling and Cross-Dataset Constraints}
To evaluate scaling behavior, we test \textsc{SYNAPSE} on the \textsc{THINGS EEG2} dataset (Table~\ref{tab:things_metrics}). Scaling from 40 to 1,654 semantic concepts increases density in the continuous feature space, which may increase latent cross-talk at the connectionist frontend. This helps explain the lower absolute metrics relative to ImageNet-EEG. The cross-exemplar protocol evaluates unseen images of seen concepts, so these results should not be interpreted as concept-disjoint generalization.

Despite this dataset-level performance drop, \textsc{SYNAPSE} improves some decoder settings under high-density retrieval conditions.under high-density noise. Under the unaugmented \textsc{SENSE} baseline, \texttt{Qwen2.5-7B} reaches $9.4$ ROUGE-1 and $2.0$ fluency, while the full \textsc{SYNAPSE} configuration reaches $21.8$ ROUGE-1 and $4.2$ fluency. At the same time, adequacy remains uniformly low at $1.0$ across THINGS EEG2 configurations, limiting strong claims about semantic correctness.

Finally, the experiment shows a key trade-off concerning the cross-modal syntactic exemplar matrix ($\mathbf{\mathcal{E}}_{\text{exemplars}}$). For compact open-weights architectures like \texttt{Meta-LLaMA-3-8B}, removing exemplars drops ROUGE-1 from $20.3$ to $15.2$, suggesting that smaller models may rely on localized contextual patterns to structure output text. For higher-capacity models like \texttt{GPT-4o-mini}, the small performance gap suggests that nearest-neighbor exemplars can sometimes introduce stylistic constraints. This indicates that future scaling may require dynamic retrieval or thresholding strategies to mitigate coordinate interference (see Appendix \ref{sec:appendix_prompts} for qualitative prompt details).
\section{Conclusion}
We introduced \textsc{SYNAPSE}, a lightweight neuro-symbolic framework for EEG-to-text decoding that stabilizes filters EEG-derived candidates before frozen language model decoding through inference-time symbolic regularization rather than resource-intensive end-to-end fine-tuning. By combining topological graph filtering, relational grounding, and latent exemplar retrieval, \textsc{SYNAPSE} reduces the effect of noisy neural projections on generation. Experiments across two EEG benchmarks and multiple LLM backends show improvements in several settings, limited degradation under object-label ablation, and competitive performance relative to selected fine-tuned systems. More broadly, our findings suggest that brain-to-text decoding may benefit from a retrieval-augmented approach, where symbolic structure and external semantic memory can support generation without modifying model parameters. By externalizing semantic correction away from autoregressive weights and into structured inference-time retrieval, \textsc{SYNAPSE} offers a scalable, privacy-conscious, and computationally efficient direction for next-generation neuro-symbolic brain-computer interfaces.
\section{Limitations}
Despite the robustness gains introduced by \textsc{SYNAPSE}, several limitations remain. Non-invasive EEG recordings possess inherently low spatial resolution and remain highly susceptible to physiological noise, placing fundamental constraints on the fidelity of recovered semantic intent. While topological graph purification substantially reduces semantic drift and attentional dispersion, the framework still assumes that graph connectivity implies contextual validity; consequently, semantically incorrect but densely co-activated token clusters may occasionally evade pruning and propagate residual hallucinations into downstream generation. In addition, the effectiveness of symbolic regularization depends on the coverage and relational completeness of the external commonsense graph, which may omit rare or weakly connected visual concepts. The exemplar retrieval module is similarly constrained by static nearest-neighbor matching within dense latent neural embedding spaces, where semantically adjacent concepts can overlap under large-scale open-vocabulary settings. More broadly, \textsc{SYNAPSE} inherits a limitation common to retrieval-augmented paradigms: downstream generation quality remains fundamentally bounded by retrieval quality, which in our setting operates over noisy neural semantic projections rather than clean textual queries. Finally, although raw EEG processing and symbolic purification remain fully localized to preserve biometric privacy, several evaluated decoder backends rely on externally hosted LLM APIs, meaning full end-to-end privacy ultimately depends on deployment with fully local language models and on-device inference infrastructure.

\section{Ethics Statement}
\textsc{SYNAPSE} is designed as a lightweight and privacy-conscious framework for EEG-to-text decoding, with the goal of supporting assistive communication technologies for individuals affected by neurological impairments. The architecture performs neural feature extraction and symbolic regularization entirely within the local device boundary, ensuring that raw brain signals are never transmitted to external inference services and thereby reducing risks associated with sensitive biometric exposure. To minimize ungrounded or hallucinated text generation, \textsc{SYNAPSE} incorporates explicit semantic regularization through topological graph purification and constrained vocabulary grounding. Additionally, the framework operates using generalized neural alignment representations without requiring subject-specific fine-tuning, promoting broader accessibility and reducing dependence on individualized calibration. We emphasize user privacy, transparent inference, and responsible deployment throughout the design of the system. We utilize the ConceptNet knowledge graph under its Creative Commons Attribution-ShareAlike 4.0 license and adhere to the usage terms for the publicly available EEG benchmarks used in our training pipeline.

\section{Potential Risks}
While \textsc{SYNAPSE} is developed strictly for assistive communication, we acknowledge the inherent dual-use nature of brain-computer interface (BCI) technologies. The ability to decode neural activity into semantic concepts could theoretically be repurposed for non-consensual cognitive monitoring or unauthorized neural profiling. To mitigate these risks, our research framework intentionally constrains the decoding process to perceived visual stimuli rather than internal thoughts or emotive states. Furthermore, by strictly localizing all neural processing to the user's device—thereby eliminating the need to transmit raw brain data to cloud services—we prioritize individual autonomy and data sovereignty as fundamental safeguards against the potential misuse of neural decoding technology.

\section*{Generative AI Disclosure}
Generative AI tools, including ChatGPT and Gemini, were used during manuscript preparation to assist with \LaTeX\ formatting, language refinement, conciseness, and debugging support. All authors carefully reviewed, edited, and take full responsibility for the final manuscript content.

\clearpage
\bibliography{custom}

\clearpage
\onecolumn
\appendix

\section{Qualitative Generation Examples}
\label{sec:appendix_qualitative}
\nopagebreak
This appendix provides a representative suite of translated natural language sentences generated from non-invasive biological signal features, demonstrating the semantic error-correction capabilities of our framework.
\begin{table*}[h]
\centering
\small
\setlength{\tabcolsep}{3pt}
\begin{tabular}
{>{\centering\arraybackslash}m{0.03\linewidth} 
 >{\centering\arraybackslash}m{0.11\linewidth} 
 >{\centering\arraybackslash}m{0.09\linewidth} 
 >{\centering\arraybackslash}m{0.09\linewidth} 
 >{\centering\arraybackslash}m{0.15\linewidth} 
 >{\centering\arraybackslash}m{0.15\linewidth} 
 >{\centering\arraybackslash}m{0.12\linewidth} 
 >{\centering\arraybackslash}m{0.17\linewidth}} 
\toprule
\textbf{ID} & \textbf{Image} & \textbf{Ref. Obj} & \textbf{Pred. Obj} & \textbf{Reference Description} & \textbf{Thought2Text} & \textbf{Pruned BoW} & \textbf{SYNAPSE Description} \\
\midrule
1 &
\includegraphics[width=0.10\textwidth]{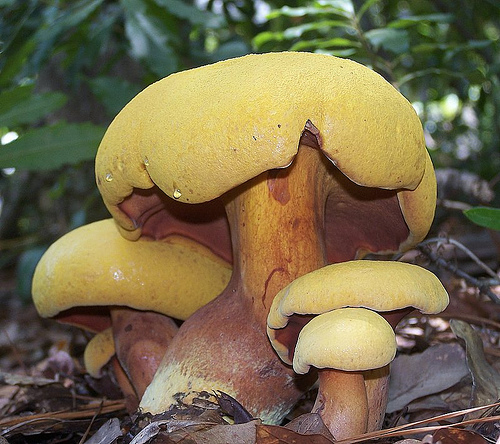} & mushroom & mushroom &
A large yellow mushroom with a brown stem and a brown cap, surrounded by green foliage. &
A group of mushrooms growing on a log. &
yellow, mushroom, white, daisy, flower, grow, green forest &
\textcolor{OliveGreen}{A yellow mushroom growing in a green forest.} \\
\midrule

2 &
\includegraphics[width=0.10\textwidth]{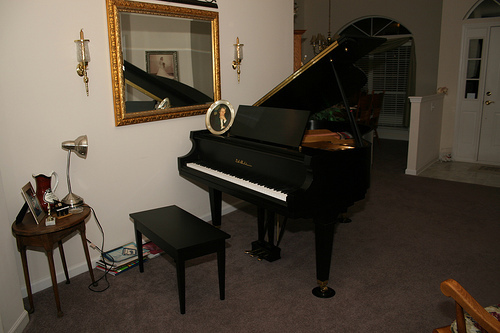} &
piano &
piano &
A black grand piano in a living room. &
A grand piano with a stool in front of it. &
piano, black, room, grand, wooden, floor &
\textcolor{OliveGreen}{A black grand piano in a room with a wooden floor and chair.} \\
\midrule

3 &
\includegraphics[width=0.10\textwidth]{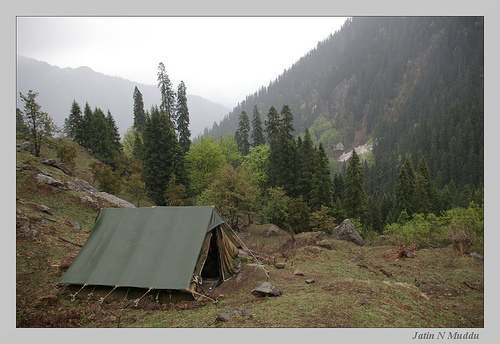} &
camping &
camping &
A tent in a mountainous area with trees and fog. &
A tent set up in a forest with a campfire nearby. &
tent, white, blue, mountains, field, area &
\textcolor{OliveGreen}{Two green tents in a grassy field with mountains in the background.} \\
\midrule

4 &
\includegraphics[width=0.10\textwidth]{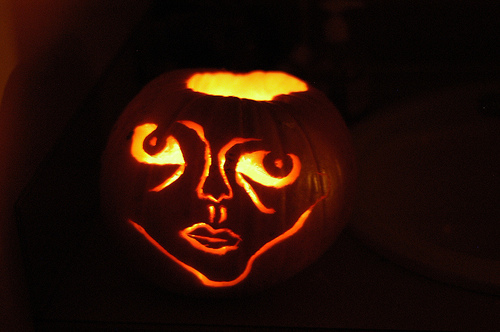} &
pumpkin &
pumpkin &
A carved pumpkin with a face and eyes, sitting on a table. &
A carved pumpkin with a spooky face on it. &
carved, pumpkin, face, carving, spooky &
\textcolor{OliveGreen}{A carved pumpkin with a jack-o-lantern face.} \\
\midrule

5 &
\includegraphics[width=0.10\textwidth]{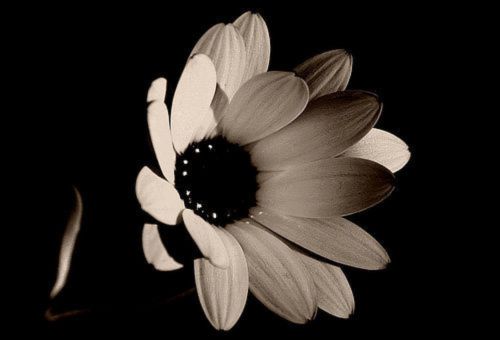} &
flower &
mushroom &
A black and white photograph of a single daisy with a white center and a dark brown center. &
A group of mushrooms growing on a log. &
pizza, mushroom, pepperoni, cheese, white, yellow &
\textcolor{BrickRed}{A pizza with mushrooms and cheese on a white table.} \\
\midrule

6 &
\includegraphics[width=0.10\textwidth]{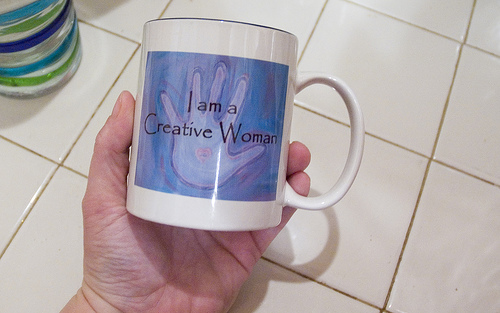} &
coffee mug &
coffee mug &
A hand holding a mug with a blue background and a handprint design. &
A person holding a coffee mug with the words ``World's Best Dad'' written on it. &
car, convertible, mug, green, coffee, vehicle &
\textcolor{BrickRed}{A green convertible driving on the street with a white coffee mug on the dashboard.} \\
\midrule

7 &
\includegraphics[width=0.10\textwidth]{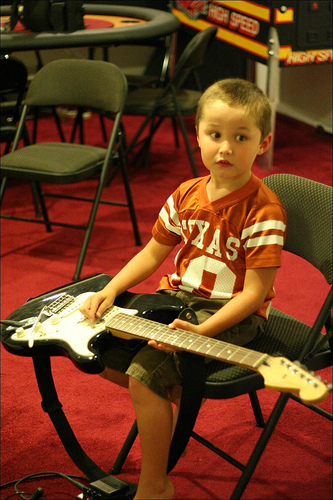} &
guitar &
train &
A young boy sitting on a chair playing a guitar. &
A man is holding a guitar in front of a microphone. &
train, track, black, blue, white, red, station &
\textcolor{BrickRed}{A train on a track at a station, red and black.} \\
\midrule

8 &
\includegraphics[width=0.10\textwidth]{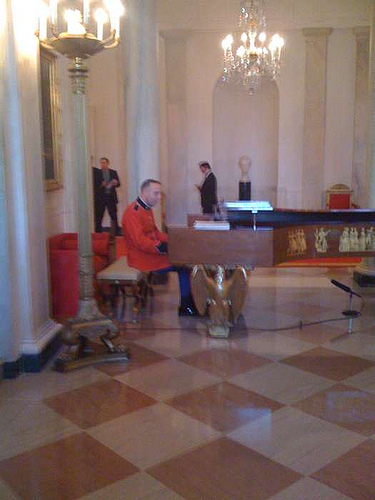} &
piano &
ball &
A man in a red coat and black pants is playing a piano in a room with a chandelier. &
A man is playing the piano in a dimly lit room. &
piano, golf, ball, black, room, white, floor, orchestra &
\textcolor{BrickRed}{A man standing in a room with black floor, playing a piano at an orchestra while golf balls are nearby.} \\

\bottomrule
\end{tabular}

\caption{Qualitative comparison of generated descriptions from \textsc{SYNAPSE} and the \textsc{Thought2Text} baseline. The Pruned BoW column details the resultant pruned token footprint after the topological graph refining phase is completed, \textit{captions generated by Qwen2.5-7B}.}
\label{tab:qualitative}
\end{table*}

\clearpage
\section{Inference-Time Prompt Blueprints}
\label{sec:appendix_prompts}
To guarantee reproducibility across inference endpoints, we detail the prompt layouts generated by the \textsc{SYNAPSE} framework. 

\subsection{ImageNet Evaluation Profiles}

\subsubsection{Main Production Architecture (Full Graph RAG Augmentation)}
This blueprint represents configuration $\mathcal{A}_1$, delivering the full consolidated neuro-symbolic payload—including topologically pruned vocabulary vectors $\mathbf{W}_{\text{pruned}}$, grounded relational knowledge blocks $\mathbf{\mathcal{F}}$, cross-modal syntactic templates $\mathbf{\mathcal{E}}_{\text{exemplars}}$, and continuous perceptual classification metrics—to the language decoder.

\begin{tcolorbox}[
  colback=gray!6,
  colframe=gray!45,
  boxrule=0.4pt,
  arc=1mm,
  left=4pt,
  right=4pt,
  top=4pt,
  bottom=4pt
]
\small
\begin{verbatim}
You are an advanced neural decoding translation engine. You are provided with a denoised, 
common-sense validated Bag-of-Words extracted from human EEG signals, along with topological 
world constraints and structural sentence guidelines. Your task is to synthesize these 
primitives into a single natural description.

[Structural Layout Guides from Training Set]
{retrieved_exemplars}

[Denoised Brain-Signal Keywords]
[{prompt_words}]

[Topological Common-Sense Relations]
{relational_facts}

[Target Dominant Signal Context]
- Primary Classification Target: '{pred_obj}' (Model Confidence: {pred_conf})

Instructions:
1. Synthesize these primitives into exactly ONE clear, fluent English description 
    (8-20 words).
2. Prioritize concepts verified by both the brain-signal keywords and the common-sense 
    relational constraints.
3. Do NOT include annotations, prefix strings, quotes, or conversational meta-commentary. 
   Output ONLY the raw caption string text.

Output:
\end{verbatim}
\end{tcolorbox}

\subsubsection{Graph Cleaned Bag-of-Words Minimal Baseline ($\mathcal{A}_6$)}
This configuration isolates the base limits of our network-filtering mechanism, conditioning the decoder exclusively on the degree centrality filtered keywords alongside standard classification arrays.

\begin{tcolorbox}[
  colback=gray!6,
  colframe=gray!45,
  boxrule=0.4pt,
  arc=1mm,
  left=4pt,
  right=4pt,
  top=4pt,
  bottom=4pt
]
\small
\begin{verbatim}
You are an advanced neural decoding translation engine. You are provided with a denoised, 
common-sense validated Bag-of-Words extracted from human EEG signals. Your task is to 
synthesize these primitives into a single natural description.

[Denoised Brain-Signal Keywords]
[{prompt_words}]

[Target Dominant Signal Context]
- Primary Classification Target: '{pred_obj}' (Model Confidence: {pred_conf})

Instructions:
1. Synthesize these primitives into exactly ONE clear, fluent English description 
    (8-20 words).
2. Prioritize concepts verified by the brain-signal keywords.
3. Do NOT include annotations, prefix strings, quotes, or conversational meta-commentary. 
   Output ONLY the raw caption string text.

Output:
\end{verbatim}
\end{tcolorbox}

\subsubsection{Context Resilience Prompt Track ($\mathcal{A}_3$)}
Designed to measure framework vulnerability to isolated connectionist failures, this configuration decouples the generation sequence from primary perceptual tracking classes, relying on the residual symbolic landscape.

\begin{tcolorbox}[
  colback=gray!6,
  colframe=gray!45,
  boxrule=0.4pt,
  arc=1mm,
  left=4pt,
  right=4pt,
  top=4pt,
  bottom=4pt
]
\small
\begin{verbatim}
You are an advanced neural decoding translation engine. You are provided with a denoised, 
common-sense validated Bag-of-Words extracted from human EEG signals, along with topological 
world constraints and structural sentence guidelines. Your task is to synthesize these 
primitives into a single natural description.

[Structural Layout Guides from Training Set]
{retrieved_exemplars}

[Denoised Brain-Signal Keywords]
[{prompt_words}]

[Topological Common-Sense Relations]
{relational_facts}

Instructions:
1. Synthesize these primitives into exactly ONE clear, fluent English description 
    (8-20 words).
2. Prioritize concepts verified by both the brain-signal keywords and the common-sense
    relational constraints.
3. Do NOT include annotations, prefix strings, quotes, or conversational meta-commentary. 
   Output ONLY the raw caption string text.

Output:
\end{verbatim}
\end{tcolorbox}

\subsubsection{Relational Fact Ablation Configuration ($\mathcal{A}_5$)}
This template deactivates the relational background constraints cache $\mathbf{\mathcal{F}}$. It forces the autoregressive layers to process structural layout guides and keyword vectors without explicit common-sense grounding linkages.

\begin{tcolorbox}[
  colback=gray!6,
  colframe=gray!45,
  boxrule=0.4pt,
  arc=1mm,
  left=4pt,
  right=4pt,
  top=4pt,
  bottom=4pt
]
\small
\begin{verbatim}
You are an advanced neural decoding translation engine. You are provided with a denoised, 
common-sense validated Bag-of-Words extracted from human EEG signals, along with structural 
sentence guidelines. Your task is to synthesize these primitives into a single 
natural description.

[Denoised Brain-Signal Keywords]
[{prompt_words}]

[Structural Layout Guides from Training Set]
{retrieved_exemplars}

[Target Dominant Signal Context]
- Primary Classification Target: '{pred_obj}' (Model Confidence: {pred_conf})

Instructions:
1. Synthesize these primitives into exactly ONE clear, fluent English description 
    (8-20 words).
2. Prioritize concepts verified by both the brain-signal keywords and the common-sense
    relational constraints.
3. Do NOT include annotations, prefix strings, quotes, or conversational meta-commentary. 
   Output ONLY the raw caption string text.

Output:
\end{verbatim}
\end{tcolorbox}

\subsubsection{In-Context Exemplar Blueprint Ablation Configuration ($\mathcal{A}_4$)}
This setting isolates model reactions to a strict zero-shot operational setting by dropping the grammatical blueprint matrix $\mathbf{\mathcal{E}}_{\text{exemplars}}$ while preserving connectionist and relational anchors.

\begin{tcolorbox}[
  colback=gray!6,
  colframe=gray!45,
  boxrule=0.4pt,
  arc=1mm,
  left=4pt,
  right=4pt,
  top=4pt,
  bottom=4pt
]
\small
\begin{verbatim}
You are an advanced neural decoding translation engine. You are provided with a denoised, 
common-sense validated Bag-of-Words extracted from human EEG signals, along with topological 
world constraints. Your task is to synthesize these primitives into a single 
natural description.

[Denoised Brain-Signal Keywords]
[{prompt_words}]

[Topological Common-Sense Relations]
{relational_facts}

[Target Dominant Signal Context]
- Primary Classification Target: '{pred_obj}' (Model Confidence: {pred_conf})

Instructions:
1. Synthesize these primitives into exactly ONE clear, fluent English description 
    (8-20 words).
2. Prioritize concepts verified by both the brain-signal keywords and the common-sense 
    relational constraints.
3. Do NOT include annotations, prefix strings, quotes, or conversational meta-commentary. 
   Output ONLY the raw caption string text.

Output:
\end{verbatim}
\end{tcolorbox}

\subsection{\textsc{THINGS EEG2} Scaling Profiles}
To trace the scaling dynamics within the experimental scope, we document the prompting profiles configured for the multi-subject \textsc{THINGS EEG2} matrix.

\subsubsection{\textsc{THINGS EEG2} Full Scaling Model Architecture ($\mathcal{B}_1$)}
This prompt represents the active baseline configuration for the \textsc{THINGS EEG2} task, tracking multi-class alignments over the vast candidate universe by aggregating the filtered token matrix alongside structural nearest-neighbor training vectors.

\begin{tcolorbox}[
  colback=gray!6,
  colframe=gray!45,
  boxrule=0.4pt,
  arc=1mm,
  left=4pt,
  right=4pt,
  top=4pt,
  bottom=4pt
]
\small
\begin{verbatim}
You are an advanced neural decoding translation engine. You are provided with a denoised, 
common-sense validated Bag-of-Words extracted from human EEG signals, along with topological 
world constraints and exemplar captions from similar samples from training data. Your task is 
to synthesize these primitives into a single natural description.

[Denoised Brain-Signal Keywords]
[{words_str}]

[Topological Common-Sense Relations]
{facts_str}

[Retrieved Exemplars]
{exemplars_str}

Instructions:
1. Synthesize these primitives into exactly ONE clear, fluent English description 
    (8-20 words).
2. Prioritize concepts verified by both the brain-signal keywords and the common-sense 
    relational constraints.
3. Do NOT include annotations, prefix strings, quotes, or conversational meta-commentary. 
   Output ONLY the raw caption string text.

Output:
\end{verbatim}
\end{tcolorbox}

\subsubsection{\textsc{THINGS EEG2} Syntactic Blueprint Matrix Ablation ($\mathcal{B}_2$)}
This template strips away cross-modal target coordinates ($\mathbf{\mathcal{E}}_{\text{exemplars}}$), forcing the autoregressive decoder to structure descriptions under high crowding metrics without localized sentence templates.

\begin{tcolorbox}[
  colback=gray!6,
  colframe=gray!45,
  boxrule=0.4pt,
  arc=1mm,
  left=4pt,
  right=4pt,
  top=4pt,
  bottom=4pt
]
\small
\begin{verbatim}
You are an advanced neural decoding translation engine. You are provided with a denoised, 
common-sense validated Bag-of-Words extracted from human EEG signals, along with topological 
world constraints. Your task is to synthesize these primitives into a single natural 
description.

[Denoised Brain-Signal Keywords]
[{words_str}]

[Topological Common-Sense Relations]
{facts_str}

Instructions:
1. Synthesize these primitives into exactly ONE clear, fluent English description 
    (8-20 words).
2. Prioritize concepts verified by both the brain-signal keywords and the common-sense 
    relational constraints.
3. Do NOT include annotations, prefix strings, quotes, or conversational meta-commentary. 
   Output ONLY the raw caption string text.

Output:
\end{verbatim}
\end{tcolorbox}

\end{document}